%% file: main.tex
\documentclass{article}
\usepackage{iclr2027_conference,times}
\usepackage[T1]{fontenc}
\usepackage[utf8]{inputenc}
\DeclareUnicodeCharacter{2212}{\ensuremath{-}}
\usepackage{amsmath,amssymb}
\usepackage{graphicx}
\usepackage{capt-of}
\usepackage{booktabs,tabularx,array}
\usepackage{microtype}
\usepackage{multirow}
\usepackage{hyperref}
\usepackage{url}

\hypersetup{
  hidelinks,
  pdftitle={MV-JEPA: Post-Training Video Encoders for Cross-View Understanding},
  pdfauthor={Yangchen Zhao, Ali Payani, Caiwen Ding}
}

\title{MV-JEPA: Post-Training Video Encoders for Cross-View Understanding}

\author{
Yangchen Zhao$^{1}$ \quad
Ali Payani$^{2}$ \quad
Caiwen Ding$^{1}$\\
$^{1}$University of Minnesota, Twin Cities \quad
$^{2}$Cisco
}

\iclrfinalcopy

\begin{document}
\maketitle

% Clear the conference-publication header for the arXiv version.
\lhead{}

\begin{abstract}
\input{sections/abstract}

\end{abstract}

\input{sections/introduction}
\input{sections/related_work}
\input{sections/method}
\input{sections/experiments}
\input{sections/conclusion}
\label{page:last-main}

% The official initial-submission limit is nine pages of main text.
% AI disclosure and references are outside that limit.
\clearpage
\input{sections/ai_use}
\bibliography{references}
\bibliographystyle{iclr2027_conference}

\clearpage
\appendix
\end{document}

%% file: sections/abstract.tex
JEPA-based video encoders learn strong representations for video and image understanding through prediction in latent space. However, their standard pretraining predicts features within individual images or video clips, without explicitly exploiting paired camera views of the same object or moment. Consequently, the learned representations may not reliably support matching across different camera viewpoints.
To solve this problem, we introduce multi-view predictive post-training to strengthen cross-view instance and temporal correspondence while retaining independent single-view inference. A shared encoder processes masked context views independently, and a training-time predictor jointly predicts latent tokens for masked context regions and a fully held-out view. We alternate cross-view prediction with single-view masked prediction on general image and video data to help retain the encoder’s pretrained capabilities. After only 9,000 post-training updates on a single GPU, MV-JEPA improves cross-view
instance retrieval and coarse cross-camera moment retrieval on various datasets, with further gains in zero-shot action-phase retrieval
and frozen-feature classification on AE2. Frozen-probe evaluations further indicate that post-training largely preserves motion recognition, image recognition, and dense semantic segmentation performance on the evaluated
benchmarks. Together, these results demonstrate that multi-view predictive post-training can strengthen instance and coarse temporal correspondence in JEPA video encoders without modifying the encoder architecture or single-view inference interface.

%% file: sections/introduction.tex
\section{Introduction}
\label{sec:introduction}

Video encoders based on the Joint-Embedding Predictive Architecture
(JEPA) learn visual representations through large-scale self-supervised
learning~\citep{bardes2024vjepa}. V-JEPA~2, for example, achieves strong visual recognition performance
using frozen encoder features~\citep{assran2025vjepa2}. However, its pretraining objective predicts
latent features within individual images or video clips, without
explicitly exploiting paired camera views of the same object or moment.
We therefore study short multi-view post-training to strengthen
cross-view correspondence while preserving the general visual
capabilities acquired during pretraining.

Cross-view correspondence requires representations to preserve identity despite changes in appearance. Different cameras can observe the same object from opposite sides or record the same activity at the same time, yet produce substantially different images. Useful features should match these observations while distinguishing other instances or moments. We therefore focus on viewpoint robustness for instance and temporal correspondence, with these distinctions retained in independently encoded features.

Prior work has explored cross-view representation learning through both
contrastive and predictive objectives.
Time-Contrastive Networks adapt an ImageNet-initialized visual model using
synchronized multi-view contrastive learning~\citep{sermanet2017tcn},
but their evaluation focuses on correspondence, task-specific attributes,
and imitation without examining recognition retention after adaptation.
BYOV combines self-view and cross-view masked prediction to learn temporal
representations on top of a frozen pretrained image encoder~\citep{park2025byov}, leaving the pretrained visual backbone itself
unchanged during cross-view training.
Seq-JEPA learns invariant and equivariant representations through
transformation-conditioned sequential prediction~\citep{ghaemi2025seqjepa}.
Its experiments study self-supervised representation learning from scratch,
without examining post-training adaptation of an already pretrained
backbone.
Its original formulation also uses explicit relative transformation
parameters as conditioning inputs during training.
Building on these advances, we study whether short post-training
can strengthen cross-view correspondence in a pretrained
JEPA video encoder while preserving its general recognition
capabilities.

We introduce MV-JEPA, a multi-view predictive post-training
recipe that directly updates the pretrained encoder.
A shared encoder independently processes masked context views,
and a joint predictor reconstructs masked context features
and predicts the features of a fully held-out view.
The held-out observation is used only to construct training
targets, requiring its features to be inferred from other views
without explicit inter-view transformation parameters.
We further alternate this objective with single-view masked-prediction
replay to help retain pretrained recognition capabilities.

Our contributions are:
\begin{enumerate}
    \item \textbf{Short multi-view predictive post-training.}
    MV-JEPA strengthens pretrained encoders with only
    9,000 updates on a single GPU, preserving their
    original single-view inference interface.

    \item \textbf{Transferable cross-view understanding.}
    Improved instance and moment correspondence transfers
    across datasets and JEPA backbones, with additional
    benefits for ego--exo action understanding.

    \item \textbf{Preserved general visual capabilities.}
    Cross-view gains coexist with largely retained motion
    recognition, image recognition, and dense semantic
    segmentation performance.
\end{enumerate}

After 9,000 post-training updates, MV-JEPA (V-JEPA2-L)
improves ABO instance retrieval by 9.4 percentage points
and moment retrieval on unseen Ego-Exo4D captures by 14.6 points.
The gains transfer to datasets not used for post-training
and to V-JEPA2.1-g~\citep{murlabadia2026vjepa21}.
On AE2~\citep{xue2023ae2}, zero-shot retrieval improves
across all four actions, while classification and temporal
ordering improve on three.
Frozen probes show largely retained recognition performance
on SSv2, Diving48, Jester, and ImageNet-1K, as well as
comparable dense semantic segmentation performance on ADE20K.

%% file: sections/related_work.tex
\section{Related Work}\label{related-work}

\textbf{Cross-view correspondence learning.}
Prior work has studied correspondence through metric learning, temporal
alignment, and predictive objectives.
TCN learns embeddings from synchronized camera views and temporal negatives~\citep{sermanet2017tcn}, while TCC learns temporal alignment through
cycle consistency~\citep{dwibedi2019tcc}.
AE2 aligns unpaired ego--exo videos through object-centric
encoding and a contrastive temporal-alignment objective~\citep{xue2023ae2}.
Among predictive approaches, CroCo reconstructs masked pixels from visible image regions and an additional view of the same scene~\citep{weinzaepfel2022croco}.
BYOV combines self-view and cross-view prediction of frame embeddings
on top of a frozen image encoder \citep{park2025byov}.
Seq-JEPA studies transformation-conditioned sequential prediction for
invariant and equivariant representations in a from-scratch training
setting \citep{ghaemi2025seqjepa}.
Our recipe directly updates a pretrained JEPA video backbone through
joint prediction of masked context regions and a held-out view, while
retaining independent single-view inference.

\textbf{Post-training of visual representations.}
Targeted additional training can improve the capabilities of pretrained
visual encoders.
TimeTuning adapts image representations using temporal-alignment clustering
on unlabeled videos \citep{salehi2023timetuning}, and NeCo improves dense
features through patch-neighbor ordering consistency
\citep{pariza2025neco}.
MoSiC uses tracked motion trajectories to enforce temporal cluster
consistency \citep{salehi2025mosic}, while 3DPoV uses tracked patches and
ordering consistency to improve dense correspondence and geometric features
\citep{simion2026_3dpov}.
Adaptation also raises questions of retention: WiSE-FT interpolates
pretrained and fine-tuned weights to balance adaptation with robustness
under distribution shift \citep{wortsman2022wiseft}.
MoSiC additionally measures image classification performance alongside
dense-task gains \citep{salehi2025mosic}.
We study this adaptation problem in JEPA video encoders through masked-context
and held-out-view latent prediction, interleaved with single-view replay,
and evaluate instance and moment correspondence alongside recognition
retention.

%% file: sections/method.tex
\section{Multi-view Predictive Post-training}
\label{method}

\begin{figure}[t]
\centering
\includegraphics[width=\linewidth]{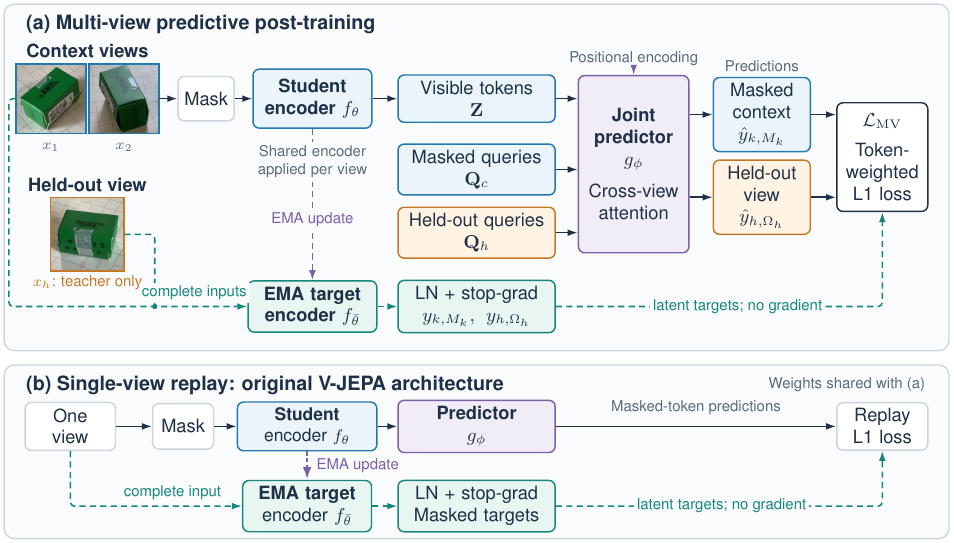}
\caption{MV-JEPA overview.
(a) A joint predictor reconstructs masked context features
and predicts held-out-view features from independently
encoded context views.
(b) Single-view replay follows the original V-JEPA
masked-prediction architecture and shares encoder and
predictor weights with the multi-view branch.
Each epoch contains 152 multi-view updates and
148 replay updates.}
\label{fig:architecture}
\end{figure}

MV-JEPA adapts pretrained video encoders for cross-view
understanding through short multi-view predictive post-training.
It combines prediction of a held-out view from multiple context
views with single-view replay of the original masked-prediction
task.
Figure~\ref{fig:architecture} illustrates the framework.

\paragraph{Training data and preprocessing.}
We use two multi-view datasets for post-training.
MVImgNet2.0 \citep{han2024mvimgnet2} contains images of everyday objects captured
from different viewpoints, extracted from short videos
recorded around each object.
Ego-Exo4D~\citep{grauman2024egoexo4d} contains synchronized egocentric and exocentric
recordings of skilled human activities, including cooking,
sports, and music.

We sample from the two datasets with equal probability.
Each training example consists of $K$ context views,
$\{x_k\}_{k=1}^{K}$, and one additional held-out view $x_h$.
For MVImgNet2.0, the context comprises two to six images
of the same object from different viewpoints.
For Ego-Exo4D, it comprises up to three synchronized clips
covering the same temporal interval from different viewpoints.

For MVImgNet2.0, each image is preprocessed using a square crop
derived from its foreground mask.
A target foreground-area ratio in $[0.35,0.55]$ is sampled
once per instance and shared across its views.
Crop sizes adapt to each view while preserving the full
foreground bounding box and retaining local background.
The crops are resized to the input resolution
(Figure~\ref{fig:mvimgnet_preprocessing}), emphasizing the
shared object in multi-view prediction.
All matched training controls use the same preprocessed data.

\input{figures/mvimgnet_preprocessing}

\paragraph{Context encoding and latent targets.}
Our main implementation initializes the encoder and predictor
from released V-JEPA2-L weights and retains its masked encoding
and EMA target construction \citep{assran2025vjepa2}.
As illustrated in Figure~\ref{fig:architecture},
$f_\theta$ denotes the student encoder and
$f_{\bar\theta}$ its EMA target encoder.

For context view $x_k$, let $V_k$ and $M_k$ denote
the visible and masked token positions, respectively.
The student independently encodes each view's visible tokens
to obtain context features $z_k$.
The teacher processes the complete context and held-out views
to produce target features $y_k$ and $y_h$:
\begin{equation}
\label{eq:targets}
\begin{aligned}
z_k &= f_\theta(x_k;V_k), \\
y_k &= \operatorname{sg}\!\left[
    \operatorname{LN}\!\left(f_{\bar\theta}(x_k)\right)
    \right], \\
y_h &= \operatorname{sg}\!\left[
    \operatorname{LN}\!\left(f_{\bar\theta}(x_h)\right)
    \right].
\end{aligned}
\end{equation}
Here, $\operatorname{LN}$ applies LayerNorm to each token,
and $\operatorname{sg}$ stops gradients through the targets.
We supervise masked context positions $M_k$ and all held-out
positions, denoted by $\Omega_h$.

Context masking uses eight blocks per view at nominal
spatial scale 0.15, following the V-JEPA2 block-mask sampler.
The held-out observation enters only the teacher:
neither the student nor the predictor receives its pixels
or encoded features.

\paragraph{Joint multi-view prediction.}
The predictor combines the visible features from all context
views to reconstruct masked context features and predict
the held-out view.

For context view $k$, let $z_{k,p}$ denote the student feature
at visible token position $p$.
A shared projection $P$ maps these features to the predictor
width.
We use two learned mask embeddings, $m_c$ and $m_h$,
to construct queries for masked-context and held-out
prediction, respectively.
Each query requests a feature at a position where
no visual token is provided to the predictor.

We use learnable view embeddings to distinguish tokens
from different views.
For each example, we randomly assign distinct view IDs
to the context views.
Let $s_k$ denote the embedding associated with the ID
assigned to context view $k$, and let $s_h$ denote
a separate embedding reserved for the held-out view.
Visible tokens and masked queries from the same context
view share $s_k$.
We also retain the original V-JEPA positional encoding to
identify spatial and temporal locations within each view.

We denote the resulting visible-token inputs by $u_{k,p}$,
masked-context queries by $q_{k,p}$, and held-out queries
by $q_{h,p}$:
\begin{equation}
\label{eq:view_slots}
\begin{aligned}
u_{k,p} &= P(z_{k,p}) + s_k,
    && p\in V_k, \\
q_{k,p} &= m_c + s_k,
    && p\in M_k, \\
q_{h,p} &= m_h + s_h,
    && p\in\Omega_h.
\end{aligned}
\end{equation}

Let $\mathbf Z$ collect all visible-token inputs $u_{k,p}$,
$\mathbf Q_c$ all masked-context queries $q_{k,p}$,
and $\mathbf Q_h$ all held-out queries $q_{h,p}$.
The joint predictor $g_\phi$, with parameters $\phi$,
processes their concatenation through self-attention.
Let $\hat y_{k,p}$ and $\hat y_{h,p}$ denote its predicted
features at context and held-out query positions:
\begin{equation}
\label{eq:predictor}
\left(
    \{\hat y_{k,M_k}\}_{k=1}^{K},
    \hat y_{h,\Omega_h}
\right)
=
g_\phi\left(
    \operatorname{concat}
    [\mathbf Z,\mathbf Q_c,\mathbf Q_h]
\right).
\end{equation}
Here, $\hat y_{k,M_k}$ collects predictions at the masked
positions of context view $k$, while $\hat y_{h,\Omega_h}$
collects predictions at every held-out position.
Each query can attend to visible features from all context
views, allowing both prediction tasks to use cross-view
information.

\paragraph{Multi-view prediction objective.}
Using the V-JEPA2 token prediction error
$e(\hat y,y)=D^{-1}\|\hat y-y\|_1$ for $D$-dimensional
features, we supervise masked context positions and
all held-out positions:
\begin{equation}
\label{eq:loss}
\mathcal L_{\mathrm{MV}}
=
\frac{
    \lambda_c
    \sum_{k=1}^{K}\sum_{p\in M_k}
    e(\hat y_{k,p},y_{k,p})
    +
    \lambda_h
    \sum_{p\in\Omega_h}
    e(\hat y_{h,p},y_{h,p})
}{
    \lambda_c\sum_{k=1}^{K}|M_k|
    +
    \lambda_h|\Omega_h|
}.
\end{equation}
We set $\lambda_c=\lambda_h=1$, giving equal weight
to each supervised token, and average the normalized
per-example losses over the batch.

Masked-context prediction preserves the within-view
reconstruction task while allowing additional views
to provide context.
Held-out prediction requires the context representations
to support inference of another view's token-level features.

\paragraph{Replay and training configuration.}
To help retain general recognition capabilities, we interleave
multi-view updates with single-view masked-prediction replay
from ImageNet~\citep{deng2009imagenet}, Kinetics-700~\citep{carreira2019kinetics700}, and Something-Something-v2~\citep{goyal2017something,mahdisoltani2018effectiveness}.
Replay shares the encoder and predictor with the multi-view
branch and follows the original V-JEPA objective and
two mask configurations.
Each epoch contains a shuffled schedule of 152 multi-view
and 148 replay updates.
The EMA teacher is updated after each optimizer step
with momentum $0.99925$.

For V-JEPA2-L, video inputs consist of 16 sampled frames
at $256\times256$ resolution.
For image inputs, each image is resized to the same
resolution and repeated to form a 16-frame clip.
Multi-view and replay batch sizes are 48 and 24,
respectively.
We use AdamW~\citep{loshchilov2019decoupled} with weight decay 0.04 and a post-training
budget of 30 epochs, corresponding to 9,000 optimizer updates.
Post-training V-JEPA2-L takes approximately 18 hours
on a single NVIDIA RTX PRO 6000 Blackwell GPU.

\paragraph{Transfer to V-JEPA2.1-g.}
We apply the same multi-view objective to V-JEPA2.1-g,
initializing the student and EMA teacher from its released
EMA encoder weights.
During post-training, both multi-view prediction and replay
are supervised using 5,632-dimensional targets.
These targets concatenate the EMA teacher's features from
blocks 10, 20, 30, and 40, with each layer normalized separately.
Student features from the same four layers form the
predictor inputs.
For evaluation, we adopt the final-layer feature readout
used in V-JEPA2.1 \citep{murlabadia2026vjepa21},
extracting only the frozen EMA encoder's final-layer,
1,408-dimensional features.

Images use the native single-frame tokenizer, and image
and video batches are processed in separate updates.
Replay follows V-JEPA2.1 by combining masked-token
prediction with a distance-weighted L1 loss on visible
context tokens.
Within the visible-token loss, tokens closer to masked
positions receive greater weight.
This loss term is scaled by 0.7 for images and 0.5 for videos.

%% file: figures/mvimgnet_preprocessing.tex
\begin{figure}[tbp]
\centering
\includegraphics[width=0.68\linewidth]{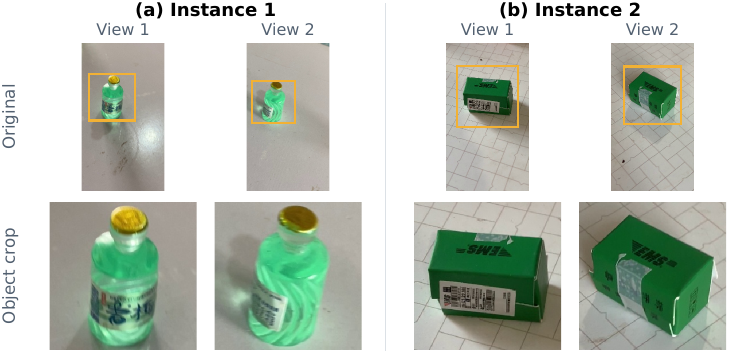}
\caption{MVImgNet2.0 preprocessing. Orange boxes mark the square crops
in the original views; the resulting $256\times256$ training
images are shown below.}
\label{fig:mvimgnet_preprocessing}
\end{figure}

%% file: sections/experiments.tex
\section{Experiments}
\label{experiments}

\paragraph{Models and shared setup.}
We primarily evaluate MV-JEPA using V-JEPA2-L.
We additionally evaluate V-JEPA2.1-g to assess transfer
of cross-view correspondence gains and retention of
semantic segmentation performance.
All post-trained models and training controls use their frozen
EMA encoders from the final epoch-30 checkpoint,
corresponding to 9,000 optimizer updates.

\subsection{Cross-view instance retrieval}
\label{cross-view-instance-retrieval}

We evaluate cross-view instance recognition using frozen encoders
and independently extracted features.
For MVImgNet2.0, we use 5,700 query--gallery pairs across 285 categories,
with 20 candidate instances per category; evaluated instances
are excluded from post-training.
For ABO~\citep{collins2022abo}, we use 7,460 spin-view pairs in 373 groups of 20 products,
with product-type-matched candidates and white square padding.
SOP~\citep{song2016deep} uses its official test set of 60,502 images and 11,316 products,
with each image querying the full gallery excluding itself.
We report Recall@1 under one fixed protocol per dataset.

\input{tables/correspondence}

As shown in Table~\ref{tab:correspondence},
MV-JEPA (V-JEPA2-L) improves Recall@1 over its released
initialization by 15.81, 9.36, and 3.64 percentage points
on MVImgNet2.0, ABO, and SOP, respectively.
MV-JEPA (V-JEPA2.1-g) achieves corresponding gains
of 11.58, 5.11, and 2.3 points.

\input{figures/abo_qualitative}

These results demonstrate improved cross-view instance discrimination,
including among objects from the same category.
Figure~\ref{fig:abo_qualitative} illustrates this ability on ABO:
MV-JEPA retrieves the same product from another viewpoint,
while the released encoder selects a different product
from the same category.

The gains on ABO and SOP show that this capability transfers
to datasets not used for post-training.
Consistent improvements for V-JEPA2-L and V-JEPA2.1-g further
support the applicability of the post-training recipe
across the two evaluated backbones.

\subsection{Cross-camera moment retrieval}
\label{cross-camera-moment-retrieval}

We evaluate cross-camera moment retrieval on two synchronized
multi-view video datasets.
Ego-Exo4D contains skilled human activities, such as sports and music,
recorded from first- and third-person perspectives.
We use four exocentric cameras for each of 81 takes from
71 captures excluded from post-training.
Assembly101~\citep{sener2022assembly101} records people assembling and disassembling toy vehicles
from multiple viewpoints.
We use 360 recordings with eight static cameras per recording;
this dataset is not used for post-training.

For each execution, we divide the valid time span into six equal
intervals and sample a 16-frame clip at 4 fps around each interval's
center, yielding six non-overlapping clips per camera.
Each clip serves as a query to retrieve its synchronized counterpart
among the six clips from another camera, using cosine similarity
between independently extracted features.
We evaluate all ordered camera pairs and compute the percentage
of correct retrievals within each take or recording.
The final Top-1 score averages these accuracies, giving equal
weight to each take or recording.

As shown in Table~\ref{tab:moment_models},
MV-JEPA (V-JEPA2-L) improves Top-1 accuracy over its released
initialization by 14.64 percentage points on Ego-Exo4D
and 8.33 points on Assembly101.
MV-JEPA (V-JEPA2.1-g) achieves corresponding gains
of 12.29 and 9.16 points.

MV-JEPA more accurately matches the same moment of an activity
across viewpoints, while distinguishing it from other moments
within the same execution.
The gains on Assembly101 show transfer to a dataset not used
for post-training, while consistent improvements on both
JEPA backbones support the transferability of our post-training
recipe.
Together, these results demonstrate improved temporal
discrimination across viewpoints, supporting stronger
cross-view understanding of dynamic scenes.

\subsection{Ego--exo action understanding on AE2}
\label{ae2-evaluation}

AE2 contains egocentric and exocentric videos of
Break Eggs, Pour Milk, Pour Liquid, and Tennis Forehand,
with frame-level action-phase annotations.
We compare frozen V-JEPA2-L and MV-JEPA (V-JEPA2-L)
using its official training and test splits.
AE2 is not used in MV-post-training.

Both encoders process 16-frame clips at $256\times256$
resolution under an offline protocol.
Frame features are extracted from the corresponding
two-frame tubelet positions, retaining the original phase labels.
Positions whose constituent frames have different labels
are excluded identically for both models during
probe training and evaluation.

Classification trains an RBF SVM with $C=1$ and
\texttt{gamma=scale}, reporting weighted F1.
Zero-shot retrieval reports mAP@10 using the AE2 implementation,
with matching phase labels defining relevance
and the query video excluded from the gallery.
Both tasks use Regular and Exo$\rightarrow$Ego settings.
Regular uses mixed ego/exo views.
Exo$\rightarrow$Ego classification trains on exocentric videos
and tests on egocentric videos; retrieval uses exocentric
queries and an egocentric gallery.
Kendall's $\tau$ evaluates temporal-order preservation
of nearest-neighbor frame matches across distinct videos.

\input{tables/ae2}

Table~\ref{tab:ae2} shows retrieval gains in both settings
across all four actions.
For Pour Milk and Pour Liquid, respectively,
Exo$\rightarrow$Ego mAP@10 increases from 53.95\% to 60.05\%
and from 53.06\% to 58.19\%, while classification F1
increases from 40.09\% to 54.14\% and from 38.71\% to 57.62\%.
Classification and Kendall's $\tau$ improve on three
of the four actions.

\input{figures/ae2_qualitative}

Figure~\ref{fig:ae2_qualitative} shows successful
Exo$\rightarrow$Ego matches on Pour Milk and Break Eggs:
MV-JEPA retrieves the query's action phase,
while the released encoder selects a different phase.
Together, these results support improved action-phase
discrimination and matching across viewpoints and executions.

\subsection{General visual capabilities with frozen probes}
\label{retention-of-general-recognition-capabilities}

We assess whether MV-post-training preserves general visual
capabilities while improving cross-view understanding.
Recognition evaluations use frozen V-JEPA2-L encoders
with attentive probes, while semantic segmentation uses
frozen V-JEPA2.1-g encoders with linear probes.

\paragraph{Motion understanding.}
We evaluate SSv2, Diving48~\citep{li2018resound},
and Jester~\citep{materzynska2019jester}.
For SSv2, we compare MV-JEPA with published reference results
reproduced in \citet[Table~4]{assran2025vjepa2}.
For Diving48 and Jester, both encoders use identical
lightweight probe protocols.
We use 4,077 Diving48 training videos and 29,641 Jester
training videos, corresponding to 27.1\% and 25.0\%
of their full training sets, respectively.
Evaluation retains the complete 1,970-video Diving48
test set and 14,787-video Jester validation set.

\input{figures/ssv2_recognition}

As shown in Figure~\ref{fig:ssv2_recognition},
MV-JEPA achieves 73.9\% Top-1 accuracy on SSv2, compared with
the published V-JEPA2-L result of 73.7\%.
We follow the V-JEPA2 frozen-probe framework with a reduced
6-configuration search, using a subset of the 20-configuration
grid used for the published baseline.

Under the matched lightweight protocols
(Table~\ref{tab:general_recognition}),
Diving48 accuracy changes from 83.30\% to 82.79\%,
while Jester accuracy increases from 96.84\% to 96.98\%.
These modest changes indicate that motion recognition
is largely retained alongside the gains in cross-view
correspondence.

\paragraph{Appearance understanding.}
We evaluate ImageNet-1K using a class-balanced subset
of 200,000 training images and the complete
50,000-image validation set.
Inputs have resolution $256\times256$,
with each image repeated to 16 frames.
Both encoders use identical probe-training settings:
15 epochs and four hyperparameter configurations.
We report the best validation accuracy among these
configurations at the final probe epoch.

Table~\ref{tab:general_recognition} shows that MV-JEPA
achieves 76.57\% Top-1 accuracy, compared with 76.74\%
for the released encoder.
This small difference indicates that appearance recognition
is also largely retained under the matched protocol.

\input{tables/general_recognition_with_ade20k}
\paragraph{Dense prediction.}
We evaluate semantic segmentation on ADE20K~\citep{zhou2017scene} using its
complete sets of 20,210 training and 2,000 validation images.
For both encoders, we train a linear probe
on frozen final-layer patch features for 40,000 updates,
using $512\times512$ training crops and identical settings
with a single fixed hyperparameter configuration.
As shown in Figure~\ref{fig:ade20k_retention},
both encoders achieve closely matched performance
throughout probe training.
At the final update, MV-JEPA achieves 47.11 mIoU,
compared with 47.23 for the released encoder.
These results indicate that dense semantic segmentation performance
is largely preserved, extending the retention evidence
to dense prediction with V-JEPA2.1-g.

\subsection{Ablations and attribution}
\label{attribution}

We compare MV-JEPA (V-JEPA2-L) with two controls
to examine the contributions of multi-view context
and held-out prediction.
All three post-trained variants include the same replay branch,
with identical replay data, masks, batch sizes, and update counts.
V-JEPA2-L + single-view post-training uses one context view
without held-out prediction.
MV-JEPA w/o held-out prediction retains multi-view context
but removes held-out queries and their supervision,
including the associated prediction computation.
All variants share the post-training data sources
and 9,000-update budget.

\input{tables/singleview}

Table~\ref{tab:singleview} shows that
V-JEPA2-L + single-view post-training does not improve
over the released encoder on any of the four tasks.
MV-JEPA w/o held-out prediction provides modest gains,
reaching 53.33\% on MVImgNet2.0, 55.79\% on ABO,
44.83\% on SOP, and 30.98\% on Ego-Exo4D.
The complete MV-JEPA model further improves these scores
by 13.06, 6.39, 2.78, and 13.77 percentage points,
respectively.

These comparisons show that
additional training on the same data sources and access
to multiple context views alone do not account for
the full gains.
The held-out prediction branch substantially improves
both recognition of the same object across views
and matching of synchronized activity moments across cameras.
This supports held-out cross-view prediction as a key
component of MV-JEPA's improved cross-view understanding.

%% file: tables/correspondence.tex
\begin{table}[tbp]
\centering

\begin{minipage}[t]{0.49\linewidth}
\vspace{0pt}
\caption{Cross-view instance retrieval with frozen encoders,
Recall@1 (\%).}
\label{tab:correspondence}
\end{minipage}
\hfill
\begin{minipage}[t]{0.49\linewidth}
\vspace{0pt}
\caption{Cross-camera moment retrieval with frozen encoders,
Top-1 accuracy (\%) over six candidate windows.}
\label{tab:moment_models}
\end{minipage}

\par\noindent
\begin{minipage}[t]{0.49\linewidth}
\vspace{0pt}
\centering
\small
\setlength{\tabcolsep}{1pt}
\begin{tabular*}{\linewidth}{@{\extracolsep{\fill}}lccc@{}}
\toprule
Frozen encoder & MVImgNet2.0 & ABO & SOP \\
\midrule
V-JEPA2-L
& 50.58 & 52.82 & 43.97 \\
\textbf{MV-JEPA (V-JEPA2-L)}
& \textbf{66.39} & \textbf{62.18} & \textbf{47.61} \\
\midrule
V-JEPA2.1-g
& 61.84 & 46.70 & 45.93 \\
\textbf{MV-JEPA (V-JEPA2.1-g)}
& \textbf{73.42} & \textbf{51.81} & \textbf{48.23} \\
\bottomrule
\end{tabular*}
\end{minipage}
\hfill
\begin{minipage}[t]{0.49\linewidth}
\vspace{0pt}
\centering
\small
\setlength{\tabcolsep}{1pt}
\begin{tabular*}{\linewidth}{@{\extracolsep{\fill}}lcc@{}}
\toprule
Frozen encoder & Ego-Exo4D & Assembly101 \\
\midrule
V-JEPA2-L
& 30.11 & 31.37 \\
\textbf{MV-JEPA (V-JEPA2-L)}
& \textbf{44.75} & \textbf{39.70} \\
\midrule
V-JEPA2.1-g
& 29.53 & 37.51 \\
\textbf{MV-JEPA (V-JEPA2.1-g)}
& \textbf{41.82} & \textbf{46.67} \\
\bottomrule
\end{tabular*}
\end{minipage}

\end{table}

%% file: figures/abo_qualitative.tex
\begin{figure}[tbp]
\centering
\includegraphics[width=\linewidth]{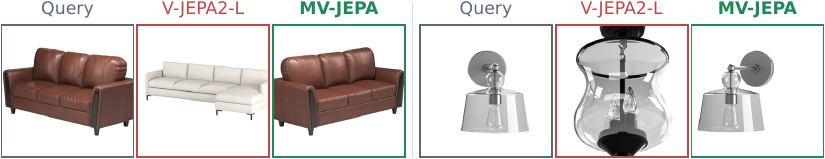}
\caption{Selected ABO retrieval examples with V-JEPA2-L
and MV-JEPA (V-JEPA2-L).
Each query uses its original gallery of 20 same-category products.
Green and red borders indicate correct and incorrect
top-1 product identities, respectively.}
\label{fig:abo_qualitative}
\end{figure}

%% file: tables/ae2.tex
\begin{table}[tbp]
\centering
\caption{AE2 action understanding with frozen encoders.
Classification and retrieval report weighted F1 (\%)
and mAP@10 (\%), respectively.}
\label{tab:ae2}
\small
\setlength{\tabcolsep}{3pt}
\begin{tabularx}{\linewidth}{
    @{}
    l
    >{\raggedright\arraybackslash}X
    *{5}{c}
    @{}
}
\toprule
& & \multicolumn{2}{c}{Classification (F1)}
& \multicolumn{2}{c}{Retrieval (mAP@10)}
& Kendall's \\
\cmidrule(lr){3-4}\cmidrule(lr){5-6}
Task & Frozen encoder
& Regular & Exo$\to$Ego
& Regular & Exo$\to$Ego
& $\tau$ $\uparrow$ \\
\midrule
\multirow{2}{*}{Break Eggs}
& V-JEPA2-L
& 63.80 & 20.17 & 59.48 & 40.93 & 0.1123 \\
& MV-JEPA (V-JEPA2-L)
& 62.70 & 19.27 & 60.07 & 43.72 & 0.1024 \\
\midrule
\multirow{2}{*}{Pour Milk}
& V-JEPA2-L
& 68.06 & 40.09 & 72.68 & 53.95 & 0.2598 \\
& MV-JEPA (V-JEPA2-L)
& 71.19 & 54.14 & 75.24 & 60.05 & 0.3154 \\
\midrule
\multirow{2}{*}{Pour Liquid}
& V-JEPA2-L
& 71.43 & 38.71 & 70.41 & 53.06 & 0.1051 \\
& MV-JEPA (V-JEPA2-L)
& 75.21 & 57.62 & 71.58 & 58.19 & 0.1093 \\
\midrule
\multirow{2}{*}{\shortstack[l]{Tennis\\Forehand}}
& V-JEPA2-L
& 92.16 & 63.79 & 90.15 & 63.53 & 0.4833 \\
& MV-JEPA (V-JEPA2-L)
& 93.68 & 66.30 & 91.74 & 63.56 & 0.5391 \\
\bottomrule
\end{tabularx}
\end{table}

%% file: figures/ae2_qualitative.tex
\begin{figure}[tbp]
\centering
\includegraphics[width=\linewidth]{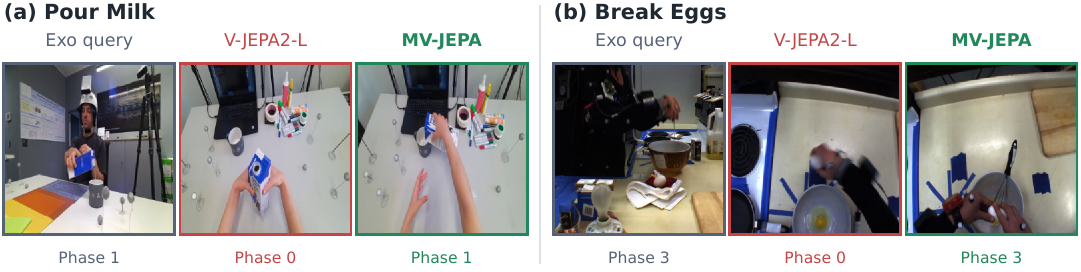}
\caption{Selected Exo$\to$Ego retrieval examples on AE2.
Each panel shows a query frame and the top-1 retrieved frames
from V-JEPA2-L and MV-JEPA.
Green and red borders indicate matching and mismatched
action phases, respectively.}
\label{fig:ae2_qualitative}
\end{figure}

%% file: figures/ssv2_recognition.tex
\begin{figure}[tbp]
\centering
\includegraphics[width=0.85\linewidth]{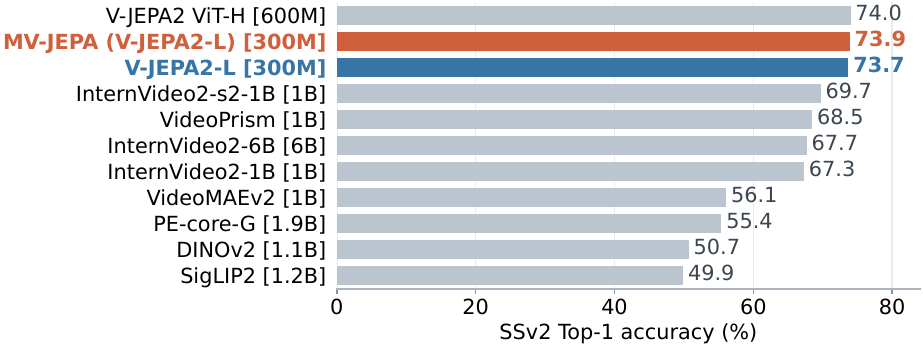}
\caption{Frozen-probe recognition on SSv2, Top-1 (\%).
Selected references include VideoMAEv2~\citep{wang2023videomaev2},
InternVideo2~\citep{wang2024internvideo2},
VideoPrism~\citep{zhao2024videoprism},
DINOv2~\citep{oquab2023dinov2},
PE~\citep{bolya2025perception},
SigLIP2~\citep{tschannen2025siglip2},
and V-JEPA2~\citep{assran2025vjepa2}; scores follow
\citet[Table~4]{assran2025vjepa2}.
MV-JEPA uses a reduced 6-configuration search compared with
the published V-JEPA2-L baseline.}
\label{fig:ssv2_recognition}
\end{figure}

%% file: tables/general_recognition_with_ade20k.tex
\begin{figure}[tbp]
\centering
\setlength{\parskip}{0pt}
% Caption and legend share a row; the table title remains above the table.
\begin{minipage}[b]{0.59\linewidth}
\captionof{table}{Frozen-probe recognition, Top-1 (\%). Both encoders use V-JEPA2-L~\citep{assran2025vjepa2} (300M) with matched lightweight protocols.}
\label{tab:general_recognition}
\end{minipage}\hfill
\begin{minipage}[b]{0.39\linewidth}
\centering
\includegraphics[width=\linewidth]{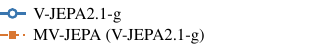}
\end{minipage}

\par\nointerlineskip\noindent
% The plot's axes height equals the table height. Keep the table's natural row spacing.
\begin{minipage}[t]{0.59\linewidth}
\vspace{3pt}
\centering
\small
\setlength{\tabcolsep}{1.5pt}
\renewcommand{\arraystretch}{1.05}
\begin{tabular*}{\linewidth}{@{\extracolsep{\fill}}lccc@{}}
\toprule
& \multicolumn{2}{c}{\shortstack[c]{Motion\\Understanding}}
& \shortstack[c]{Appearance\\Understanding} \\
\cmidrule(lr){2-3}\cmidrule(lr){4-4}
Frozen encoder & Diving48 & Jester & ImageNet-1K \\
\midrule
V-JEPA2-L & 83.30 & 96.84 & 76.74 \\
MV-JEPA (V-JEPA2-L) & 82.79 & 96.98 & 76.57 \\
\bottomrule
\end{tabular*}
\end{minipage}\hfill
\begin{minipage}[t]{0.39\linewidth}
\vspace{0pt}
\centering
\includegraphics[width=\linewidth]{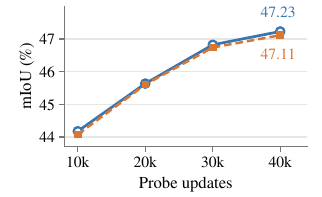}
\end{minipage}

\par\nointerlineskip\noindent\hspace*{0.61\linewidth}%
\begin{minipage}[t]{0.39\linewidth}
\captionof{figure}{ADE20K linear-probe curves.}
\label{fig:ade20k_retention}
\end{minipage}
\end{figure}

%% file: tables/singleview.tex
\begin{table}[tbp]
\centering
\caption{Ablation of multi-view post-training on V-JEPA2-L.
Scores are R@1 (\%) for instance retrieval and Top-1 (\%)
for cross-camera moment retrieval, following the protocols in
Tables~\ref{tab:correspondence} and~\ref{tab:moment_models}.}
\label{tab:singleview}
\small
\begin{tabularx}{\linewidth}{
@{}>{\raggedright\arraybackslash}Xcccc@{}
}
\toprule
Frozen encoder / variant
& MVImgNet2.0 & ABO & SOP & Ego-Exo4D \\
\midrule
V-JEPA2-L
& 50.58 & 52.82 & 43.97 & 30.11 \\
V-JEPA2-L + single-view post-training
& 50.19 & 50.63 & 43.77 & 29.97 \\
MV-JEPA w/o held-out prediction
& 53.33 & 55.79 & 44.83 & 30.98 \\
\textbf{MV-JEPA (V-JEPA2-L)}
& \textbf{66.39} & \textbf{62.18}
& \textbf{47.61} & \textbf{44.75} \\
\bottomrule
\end{tabularx}
\end{table}

%% file: sections/conclusion.tex
\section{Conclusion}
\label{conclusion}

We introduced MV-JEPA, a post-training recipe combining multi-view
latent prediction with single-view replay. After 9,000 updates, it
improves cross-view instance and coarse temporal correspondence across
datasets and two JEPA backbones, with further benefits for ego--exo
action understanding. Frozen-probe evaluations indicate largely preserved recognition and semantic segmentation performance on the evaluated benchmarks,
while matched controls support the contribution of held-out-view prediction.
These results demonstrate that targeted predictive post-training can
strengthen cross-view understanding while retaining the original
encoder architecture and independent single-view inference interface.

%% file: sections/ai_use.tex
\subsection*{AI use statement}

Generative AI tools assisted with literature search, software implementation and debugging, and manuscript polishing. The authors are responsible for the final content of this work, including AI-assisted text, claims, code, and other artifacts.

% Working disclosure: before submission, all authors should check its scope
% and document the human verification actually performed. Do not assert
% completed verification until that verification has taken place.

%% file: references.bib
@article{assran2025vjepa2,
  title = {{V-JEPA 2}: Self-Supervised Video Models Enable Understanding, Prediction and Planning},
  author = {Assran, Mahmoud and Bardes, Adrien and Fan, David and Garrido, Quentin and Howes, Russell and Komeili, Mojtaba and Muckley, Matthew and Rizvi, Ammar and Roberts, Claire and Sinha, Koustuv and Zholus, Artem and Arnaud, Sergio and Gejji, Abha and Martin, Ada and Robert Hogan, Francois and Dugas, Daniel and Bojanowski, Piotr and Khalidov, Vasil and Labatut, Patrick and Massa, Francisco and Szafraniec, Marc and Krishnakumar, Kapil and Li, Yong and Ma, Xiaodong and Chandar, Sarath and Meier, Franziska and LeCun, Yann and Rabbat, Michael and Ballas, Nicolas},
  journal = {arXiv preprint arXiv:2506.09985},
  year = {2025},
  url = {https://arxiv.org/abs/2506.09985}
}

@article{murlabadia2026vjepa21,
  title = {{V-JEPA 2.1}: Unlocking Dense Features in Video Self-Supervised Learning},
  author = {Mur-Labadia, Lorenzo and Muckley, Matthew and Bar, Amir and Assran, Mido and Sinha, Koustuv and Rabbat, Mike and LeCun, Yann and Ballas, Nicolas and Bardes, Adrien},
  journal = {arXiv preprint arXiv:2603.14482},
  year = {2026},
  url = {https://arxiv.org/abs/2603.14482}
}

@inproceedings{weinzaepfel2022croco,
  title = {{CroCo}: Self-Supervised Pre-training for {3D} Vision Tasks by Cross-View Completion},
  author = {Weinzaepfel, Philippe and Leroy, Vincent and Lucas, Thomas and Br{\'e}gier, Romain and Cabon, Yohann and Arora, Vaibhav and Antsfeld, Leonid and Chidlovskii, Boris and Csurka, Gabriela and Revaud, J{\'e}r{\^o}me},
  booktitle = {Advances in Neural Information Processing Systems},
  year = {2022},
  url = {https://arxiv.org/abs/2210.10716}
}

@article{oquab2023dinov2,
  title = {{DINOv2}: Learning Robust Visual Features without Supervision},
  author = {Oquab, Maxime and Darcet, Timoth{\'e}e and Moutakanni, Th{\'e}o and Vo, Huy and Szafraniec, Marc and Khalidov, Vasil and Fernandez, Pierre and Haziza, Daniel and Massa, Francisco and El-Nouby, Alaaeldin and Assran, Mahmoud and Ballas, Nicolas and Galuba, Wojciech and Howes, Russell and Huang, Po-Yao and Li, Shang-Wen and Misra, Ishan and Rabbat, Michael and Sharma, Vasu and Synnaeve, Gabriel and Xu, Hu and Jegou, Herv{\'e} and Mairal, Julien and Labatut, Patrick and Joulin, Armand and Bojanowski, Piotr},
  journal = {arXiv preprint arXiv:2304.07193},
  year = {2023},
  url = {https://arxiv.org/abs/2304.07193}
}

@inproceedings{sermanet2017tcn,
  title = {Time-Contrastive Networks: Self-Supervised Learning from Multi-View Observation},
  author = {Sermanet, Pierre and Lynch, Corey and Hsu, Jasmine and Levine, Sergey},
  booktitle = {Proceedings of the IEEE Conference on Computer Vision and Pattern Recognition Workshops},
  year = {2017},
  url = {https://sermanet.github.io/tcn/}
}

@inproceedings{dwibedi2019tcc,
  title = {Temporal Cycle-Consistency Learning},
  author = {Dwibedi, Debidatta and Aytar, Yusuf and Tompson, Jonathan and Sermanet, Pierre and Zisserman, Andrew},
  booktitle = {Proceedings of the IEEE/CVF Conference on Computer Vision and Pattern Recognition},
  pages = {1801--1810},
  year = {2019},
  url = {https://sites.google.com/view/temporal-cycle-consistency}
}

@inproceedings{park2025byov,
  title = {Bootstrap Your Own Views: Masked Ego-Exo Modeling for Fine-grained View-invariant Video Representations},
  author = {Park, Jungin and Lee, Jiyoung and Sohn, Kwanghoon},
  booktitle = {Proceedings of the IEEE/CVF Conference on Computer Vision and Pattern Recognition},
  year = {2025},
  url = {https://arxiv.org/abs/2503.19706}
}

@inproceedings{ghaemi2025seqjepa,
  title = {seq-{JEPA}: Autoregressive Predictive Learning of Invariant-Equivariant World Models},
  author = {Ghaemi, Hafez and Muller, Eilif Benjamin and Bakhtiari, Shahab},
  booktitle = {Advances in Neural Information Processing Systems},
  year = {2025},
  url = {https://openreview.net/forum?id=GKt3VRaCU1}
}

@article{bardes2024vjepa,
  author = {Adrien Bardes and Quentin Garrido and Jean Ponce and Xinlei Chen and Michael Rabbat and Yann LeCun and Mahmoud Assran and Nicolas Ballas},
  title = {Revisiting Feature Prediction for Learning Visual Representations from Video},
  journal = {Transactions on Machine Learning Research},
  year = {2024},
  url = {https://openreview.net/forum?id=QaCCuDfBk2}
}

@inproceedings{xue2023ae2,
  author = {Zihui Xue and Kristen Grauman},
  title = {Learning Fine-grained View-Invariant Representations from Unpaired Ego-Exo Videos via Temporal Alignment},
  booktitle = {Advances in Neural Information Processing Systems},
  volume = {36},
  year = {2023},
  url = {https://papers.nips.cc/paper_files/paper/2023/hash/a845fdc3f87751710218718adb634fe7-Abstract-Conference.html}
}

@inproceedings{salehi2023timetuning,
  author = {Mohammadreza Salehi and Efstratios Gavves and Cees G. M. Snoek and Yuki M. Asano},
  title = {Time Does Tell: Self-Supervised Time-Tuning of Dense Image Representations},
  booktitle = {Proceedings of the IEEE/CVF International Conference on Computer Vision},
  year = {2023},
  url = {https://openaccess.thecvf.com/content/ICCV2023/html/Salehi_Time_Does_Tell_Self-Supervised_Time-Tuning_of_Dense_Image_Representations_ICCV_2023_paper.html}
}

@inproceedings{pariza2025neco,
  author = {Valentinos Pariza and Mohammadreza Salehi and Gertjan J. Burghouts and Francesco Locatello and Yuki M. Asano},
  title = {Near, Far: Patch-Ordering Enhances Vision Foundation Models' Scene Understanding},
  booktitle = {International Conference on Learning Representations},
  year = {2025},
  url = {https://proceedings.iclr.cc/paper_files/paper/2025/hash/fbc9981dd6316378aee7fd5975250f21-Abstract-Conference.html}
}

@inproceedings{salehi2025mosic,
  author = {Mohammadreza Salehi and Shashanka Venkataramanan and Ioana Simion and Efstratios Gavves and Cees G. M. Snoek and Yuki M. Asano},
  title = {{MoSiC}: Optimal-Transport Motion Trajectory for Dense Self-Supervised Learning},
  booktitle = {Proceedings of the IEEE/CVF International Conference on Computer Vision},
  year = {2025},
  pages = {6541--6551},
  url = {https://openaccess.thecvf.com/content/ICCV2025/html/Salehi_MoSiC_Optimal-Transport_Motion_Trajectory_for_Dense_Self-Supervised_Learning_ICCV_2025_paper.html}
}

@inproceedings{simion2026_3dpov,
  author = {Ioana Simion and Mohammadreza Salehi and Shashanka Venkataramanan and Cees G. M. Snoek and Yuki M. Asano},
  title = {{3DPoV}: Improving 3D Understanding via Patch Ordering on Videos},
  booktitle = {Proceedings of the 43rd International Conference on Machine Learning},
  year = {2026},
  url = {https://openreview.net/forum?id=CxcHPeXYWB}
}

@inproceedings{wortsman2022wiseft,
  author = {Mitchell Wortsman and Gabriel Ilharco and Jong Wook Kim and Mike Li and Simon Kornblith and Rebecca Roelofs and Raphael Gontijo-Lopes and Hannaneh Hajishirzi and Ali Farhadi and Hongseok Namkoong and Ludwig Schmidt},
  title = {Robust Fine-Tuning of Zero-Shot Models},
  booktitle = {Proceedings of the IEEE/CVF Conference on Computer Vision and Pattern Recognition},
  year = {2022},
  url = {https://arxiv.org/abs/2109.01903}
}

@inproceedings{wang2023videomaev2,
  title = {{VideoMAE V2}: Scaling Video Masked Autoencoders with Dual Masking},
  author = {Wang, Limin and Huang, Bingkun and Zhao, Zhiyu and Tong, Zhan and He, Yinan and Wang, Yi and Wang, Yali and Qiao, Yu},
  booktitle = {Proceedings of the IEEE/CVF Conference on Computer Vision and Pattern Recognition},
  pages = {14549--14560},
  year = {2023},
  url = {https://openaccess.thecvf.com/content/CVPR2023/html/Wang_VideoMAE_V2_Scaling_Video_Masked_Autoencoders_With_Dual_Masking_CVPR_2023_paper.html}
}

@article{wang2024internvideo2,
  title = {{InternVideo2}: Scaling Foundation Models for Multimodal Video Understanding},
  author = {Wang, Yi and Li, Kunchang and Li, Xinhao and Yu, Jiashuo and He, Yinan and Wang, Chenting and Chen, Guo and Pei, Baoqi and Yan, Ziang and Zheng, Rongkun and Xu, Jilan and Wang, Zun and Shi, Yansong and Jiang, Tianxiang and Li, Songze and Zhang, Hongjie and Huang, Yifei and Qiao, Yu and Wang, Yali and Wang, Limin},
  journal = {arXiv preprint arXiv:2403.15377},
  year = {2024},
  url = {https://arxiv.org/abs/2403.15377}
}

@inproceedings{zhao2024videoprism,
  title = {{VideoPrism}: A Foundational Visual Encoder for Video Understanding},
  author = {Zhao, Long and Gundavarapu, Nitesh B. and Yuan, Liangzhe and Zhou, Hao and Yan, Shen and Sun, Jennifer J. and Friedman, Luke and Qian, Rui and Weyand, Tobias and Zhao, Yue and Hornung, Rachel and Schroff, Florian and Yang, Ming-Hsuan and Ross, David A. and Wang, Huisheng and Adam, Hartwig and Sirotenko, Mikhail and Liu, Ting and Gong, Boqing},
  booktitle = {International Conference on Machine Learning},
  year = {2024},
  url = {https://arxiv.org/abs/2402.13217}
}

@article{bolya2025perception,
  title = {{Perception Encoder}: The Best Visual Embeddings Are Not at the Output of the Network},
  author = {Bolya, Daniel and Huang, Po-Yao and Sun, Peize and Cho, Jang Hyun and Madotto, Andrea and Wei, Chen and Ma, Tengyu and Zhi, Jiale and Rajasegaran, Jathushan and Rasheed, Hanoona and Wang, Junke and Monteiro, Marco and Xu, Hu and Dong, Shiyu and Ravi, Nikhila and Li, Daniel and Doll{\'a}r, Piotr and Feichtenhofer, Christoph},
  journal = {arXiv preprint arXiv:2504.13181},
  year = {2025},
  url = {https://arxiv.org/abs/2504.13181}
}

@article{tschannen2025siglip2,
  title = {{SigLIP 2}: Multilingual Vision-Language Encoders with Improved Semantic Understanding, Localization, and Dense Features},
  author = {Tschannen, Michael and Gritsenko, Alexey and Wang, Xiao and Naeem, Muhammad Ferjad and Alabdulmohsin, Ibrahim and Parthasarathy, Nikhil and Evans, Talfan and Beyer, Lucas and Xia, Ye and Mustafa, Basil and H{\'e}naff, Olivier and Harmsen, Jeremiah and Steiner, Andreas and Zhai, Xiaohua},
  journal = {arXiv preprint arXiv:2502.14786},
  year = {2025},
  url = {https://arxiv.org/abs/2502.14786}
}

@article{han2024mvimgnet2,
  title = {{MVImgNet2.0}: A Larger-scale Dataset of Multi-view Images},
  author = {Han, Xiaoguang and Wu, Yushuang and Shi, Luyue and Liu, Haolin and Liao, Hongjie and Qiu, Lingteng and Yuan, Weihao and Gu, Xiaodong and Dong, Zilong and Cui, Shuguang},
  journal = {ACM Transactions on Graphics},
  volume = {43},
  number = {6},
  year = {2024},
  doi = {10.1145/3687973},
  url = {https://doi.org/10.1145/3687973}
}

@inproceedings{grauman2024egoexo4d,
 author = {Grauman, Kristen and Westbury, Andrew and Torresani, Lorenzo and Kitani, Kris and Malik, Jitendra and Afouras, Triantafyllos and Ashutosh, Kumar and Baiyya, Vijay and Bansal, Siddhant and Boote, Bikram and Byrne, Eugene and Chavis, Zach and Chen, Joya and Cheng, Feng and Chu, Fu-Jen and Crane, Sean and Dasgupta, Avijit and Dong, Jing and Escobar, Maria and Forigua, Cristhian and Gebreselasie, Abrham and Haresh, Sanjay and Huang, Jing and Islam, Md Mohaiminul and Jain, Suyog and Khirodkar, Rawal and Kukreja, Devansh and Liang, Kevin J and Liu, Jia-Wei and Majumder, Sagnik and Mao, Yongsen and Martin, Miguel and Mavroudi, Effrosyni and Nagarajan, Tushar and Ragusa, Francesco and Ramakrishnan, Santhosh Kumar and Seminara, Luigi and Somayazulu, Arjun and Song, Yale and Su, Shan and Xue, Zihui and Zhang, Edward and Zhang, Jinxu and Castillo, Angela and Chen, Changan and Fu, Xinzhu and Furuta, Ryosuke and Gonzalez, Cristina and Gupta, Prince and Hu, Jiabo and Huang, Yifei and Huang, Yiming and Khoo, Weslie and Kumar, Anush and Kuo, Robert and Lakhavani, Sach and Liu, Miao and Luo, Mi and Luo, Zhengyi and Meredith, Brighid and Miller, Austin and Oguntola, Oluwatumininu and Pan, Xiaqing and Peng, Penny and Pramanick, Shraman and Ramazanova, Merey and Ryan, Fiona and Shan, Wei and Somasundaram, Kiran and Song, Chenan and Southerland, Audrey and Tateno, Masatoshi and Wang, Huiyu and Wang, Yuchen and Yagi, Takuma and Yan, Mingfei and Yang, Xitong and Yu, Zecheng and Zha, Shengxin Cindy and Zhao, Chen and Zhao, Ziwei and Zhu, Zhifan and Zhuo, Jeff and Arbelaez, Pablo and Bertasius, Gedas and Damen, Dima and Engel, Jakob and Farinella, Giovanni Maria and Furnari, Antonino and Ghanem, Bernard and Hoffman, Judy and Jawahar, C.V. and Newcombe, Richard and Park, Hyun Soo and Rehg, James M. and Sato, Yoichi and Savva, Manolis and Shi, Jianbo and Shou, Mike Zheng and Wray, Michael},
  title = {{Ego-Exo4D}: Understanding Skilled Human Activity from First- and Third-Person Perspectives},
  booktitle = {Proceedings of the IEEE/CVF Conference on Computer Vision and Pattern Recognition (CVPR)},
  month = {June},
  year = {2024},
  pages = {19383--19400},
  url = {https://openaccess.thecvf.com/content/CVPR2024/html/Grauman_Ego-Exo4D_Understanding_Skilled_Human_Activity_from_First-_and_Third-Person_Perspectives_CVPR_2024_paper.html}
}

@article{carreira2019kinetics700,
  title = {A Short Note on the {Kinetics-700} Human Action Dataset},
  author = {Carreira, Joao and Noland, Eric and Hillier, Chloe and Zisserman, Andrew},
  journal = {arXiv preprint arXiv:1907.06987},
  year = {2019},
  url = {https://arxiv.org/abs/1907.06987}
}

@inproceedings{deng2009imagenet,
  title = {{ImageNet}: A Large-Scale Hierarchical Image Database},
  author = {Deng, Jia and Dong, Wei and Socher, Richard and Li, Li-Jia and Li, Kai and Fei-Fei, Li},
  booktitle = {2009 IEEE Conference on Computer Vision and Pattern Recognition},
  pages = {248--255},
  year = {2009},
  organization = {IEEE},
  url = {https://www.image-net.org/static_files/papers/imagenet_cvpr09.pdf}
}

@inproceedings{goyal2017something,
  title = {The "{Something Something}" Video Database for Learning and Evaluating Visual Common Sense},
  author = {Goyal, Raghav and Ebrahimi Kahou, Samira and Michalski, Vincent and Materzynska, Joanna and Westphal, Susanne and Kim, Heuna and Haenel, Valentin and Fruend, Ingo and Yianilos, Peter and Mueller-Freitag, Moritz and Hoppe, Florian and Thurau, Christian and Bax, Ingo and Memisevic, Roland},
  booktitle = {Proceedings of the IEEE International Conference on Computer Vision (ICCV)},
  pages = {5842--5850},
  year = {2017},
  url = {https://openaccess.thecvf.com/content_iccv_2017/html/Goyal_The_Something_Something_ICCV_2017_paper.html}
}

@article{mahdisoltani2018effectiveness,
  title = {On the Effectiveness of Task Granularity for Transfer Learning},
  author = {Mahdisoltani, Farzaneh and Berger, Guillaume and Gharbieh, Waseem and Fleet, David and Memisevic, Roland},
  year = {2018},
  journal = {arXiv preprint arXiv:1804.09235},
  doi = {10.48550/arXiv.1804.09235},
  url = {https://arxiv.org/abs/1804.09235}
}

@inproceedings{loshchilov2019decoupled,
  title = {Decoupled Weight Decay Regularization},
  author = {Loshchilov, Ilya and Hutter, Frank},
  booktitle = {International Conference on Learning Representations},
  year = {2019},
  url = {https://arxiv.org/abs/1711.05101}
}

@inproceedings{collins2022abo,
  title = {{ABO}: Dataset and Benchmarks for Real-World {3D} Object Understanding},
  author = {Collins, Jasmine and Goel, Shubham and Deng, Kenan and Luthra, Achleshwar and Xu, Leon and Gundogdu, Erhan and Zhang, Xi and Yago Vicente, Tomas F. and Dideriksen, Thomas and Arora, Himanshu and Guillaumin, Matthieu and Malik, Jitendra},
  booktitle = {Proceedings of the IEEE/CVF Conference on Computer Vision and Pattern Recognition},
  year = {2022},
  url = {https://openaccess.thecvf.com/content/CVPR2022/html/Collins_ABO_Dataset_and_Benchmarks_for_Real-World_3D_Object_Understanding_CVPR_2022_paper.html}
}

@inproceedings{song2016deep,
  title = {Deep Metric Learning via Lifted Structured Feature Embedding},
  author = {Song, Hyun Oh and Xiang, Yu and Jegelka, Stefanie and Savarese, Silvio},
  booktitle = {Proceedings of the IEEE Conference on Computer Vision and Pattern Recognition},
  year = {2016},
  url = {https://github.com/rksltnl/Deep-Metric-Learning-CVPR16}
}

@inproceedings{sener2022assembly101,
  title = {{Assembly101}: A Large-Scale Multi-View Video Dataset for Understanding Procedural Activities},
  author = {Sener, Fadime and Chatterjee, Dibyadip and Shelepov, Daniel and He, Kun and Singhania, Dipika and Wang, Robert and Yao, Angela},
  booktitle = {Proceedings of the IEEE/CVF Conference on Computer Vision and Pattern Recognition},
  year = {2022},
  url = {https://openaccess.thecvf.com/content/CVPR2022/html/Sener_Assembly101_A_Large-Scale_Multi-View_Video_Dataset_for_Understanding_Procedural_Activities_CVPR_2022_paper.html}
}

@inproceedings{li2018resound,
  title = {{RESOUND}: Towards Action Recognition without Representation Bias},
  author = {Li, Yingwei and Li, Yi and Vasconcelos, Nuno},
  booktitle = {Proceedings of the European Conference on Computer Vision},
  year = {2018},
  url = {https://openaccess.thecvf.com/content_ECCV_2018/html/Yingwei_Li_RESOUND_Towards_Action_ECCV_2018_paper.html}
}

@inproceedings{materzynska2019jester,
  title = {The {Jester} Dataset: A Large-Scale Video Dataset of Human Gestures},
  author = {Materzynska, Joanna and Berger, Guillaume and Bax, Ingo and Memisevic, Roland},
  booktitle = {Proceedings of the IEEE/CVF International Conference on Computer Vision Workshops},
  year = {2019},
  url = {https://openaccess.thecvf.com/content_ICCVW_2019/html/HANDS/Materzynska_The_Jester_Dataset_A_Large-Scale_Video_Dataset_of_Human_Gestures_ICCVW_2019_paper.html}
}

@inproceedings{zhou2017scene,
  title = {Scene Parsing Through {ADE20K} Dataset},
  author = {Zhou, Bolei and Zhao, Hang and Puig, Xavier and Fidler, Sanja and Barriuso, Adela and Torralba, Antonio},
  booktitle = {Proceedings of the IEEE Conference on Computer Vision and Pattern Recognition (CVPR)},
  year = {2017},
  url = {https://openaccess.thecvf.com/content_cvpr_2017/html/Zhou_Scene_Parsing_Through_CVPR_2017_paper.html}
}
